**inTformer: A Time-Embedded Attention-Based Transformer for Crash Likelihood Prediction at Intersections Using Connected Vehicle Data**

**B M Tazbiul Hassan Anik**
Graduate Research Assistant
University of Central Florida
4000 Central Florida Blvd Orlando FL 32816
Email: bmtazbiulhassan.anik@ucf.edu

**Zubayer Islam**
Postdoctoral Scholar
University of Central Florida
4000 Central Florida Blvd Orlando FL 32816
Email: Zubayer.Islam@ucf.edu

**Mohamed Abdel-Aty**
Pegasus Professor and Trustee Chair
Department of Civil, Environmental and Construction Engineering
University of Central Florida
4000 Central Florida Blvd Orlando FL 32816
Email: M.aty@ucf.edu

**ABSTRACT**
The real-time crash likelihood prediction model is an essential component of the proactive traffic safety management system. Over the years, numerous studies have attempted to construct a crash likelihood prediction model in order to enhance traffic safety, but mostly on freeways. In the majority of the existing studies, researchers have primarily employed a deep learning-based framework to identify crash potential. Lately, Transformer has emerged as a potential deep neural network that fundamentally operates through attention-based mechanisms. Transformer has several functional benefits over extant deep learning models such as Long Short-Term Memory (LSTM), Convolution Neural Network (CNN), etc. Firstly, Transformer can readily handle long-term dependencies in a data sequence. Secondly, Transformers can parallelly process all elements in a data sequence during training. Finally, a Transformer does not have the vanishing gradient issue. Realizing the immense possibility of Transformers, this paper proposes inTersection-Transformer (inTformer), a time-embedded attention-based Transformer model that can effectively predict intersection crash likelihood in real-time. The proposed model was evaluated using connected vehicle data extracted from INRIX and Center for Advanced Transportation Technology (CATT) Lab's Signal Analytics Platform. Acknowledging the complex traffic operation mechanism at intersection, this study developed zone-specific models by dividing the intersection region into two distinct zones: within-intersection and approach zone. The best inTformer models in 'within-intersection,' and 'approach' zone achieved a sensitivity of 73%, and 70%, respectively. The zone-level models were also compared to earlier studies on crash likelihood prediction at intersections and with several established deep learning models trained on the same connected vehicle dataset. In every scenario, this each zone-level inTformer outperformed the benchmark models confirming the viability of the proposed inTformer architecture. Furthermore, to quantify the impact of features on crash likelihood at intersections, the SHAP (SHapley Additive exPlanations) method was employed on the best performing inTformer models of 'within-intersection,' and 'approach' zone. The most critical predictors were average and maximum approach speeds, average and maximum control delays, average and maximum travel times, and percent of vehicles on green.

## INTRODUCTION

Intersections are potentially hazardous crash zones with complex traffic operating attributes. According to the Fatality Analysis (FARS) database, approximately 25% (9634) of all fatal crashes in the United States in 2021 are intersection-related. Even at the peak of Covid-19 in 2020, the share of intersection-related fatal crashes in the United States was about 25% (8824) per FARS database. Given this persistent safety-critical state of intersections, researchers have made significant efforts in the previous few decades to improve safety conditions at intersections. Though most intersection safety studies so far have focused on identifying potential features that cause crashes at intersections (Lee et al., 2017; Wu et al., 2021; Yuan & Abdel-Aty, 2018), recent advancements in proactive traffic management systems have widened the scope to improve traffic safety at intersections from multi-other dimensions (Hossain et al., 2019; Shi & Abdel-Aty, 2015).

Real-time crash likelihood prediction is an important component of proactive traffic safety management systems: it uses real-time data to anticipate crash likelihood for the next short period of time (e.g., 5, 10, 15 minutes). Numerous studies have been conducted to date in order to construct realistic and usable real-time crash prediction models for various road segments such as freeways (Abdel-Aty et al., 2023; You et al., 2017; Yu & Abdel-Aty, 2014; Q. Zheng et al., 2021), arterials (Li et al., 2020; Theofilatos, 2017; Yuan et al., 2018), and intersections (Kidando et al., 2022; Yuan et al., 2019, 2021). Unfortunately, research on crash likelihood prediction at intersections is few in comparison to other road segments. This is because the traffic environment at intersections is much more complex, and only simple traffic flow characteristics at intersections cannot capture this complication. For the effective development of real-time crash prediction models at intersections, rich and exhaustive data exhibiting traffic flow patterns at intersections are essential.

With the installation of various intelligent transportation systems, enormous real-time traffic data is now available via fixed sensors, on board sensors and automatic vehicle identification devices (M. M. Ahmed & Abdel-Aty, 2012; Li et al., 2020; Shi et al., 2016). So far, numerous types of data from various sources have been investigated with the purpose of predicting crash likelihood in real-time at intersections (Ali et al., 2023; Kidando et al., 2022; Yuan et al., 2019, 2021; L. Zheng & Sayed, 2020). The majority of the data analyzed in the existing studies came from roadside sensors. Though sensor data has enormous potential to aid in the deployment of proactive management strategies, such as real-time crash likelihood prediction models at intersections, it also has limitations. To begin, sensor data does not always provide complete coverage of intersections. Second, roadside sensors are frequently costly and laborious to set up and deploy. Thirdly, roadside sensors are susceptible to hardware failure or network outages, resulting in data loss and significant maintenance expenses. All of these sensor constraints highlight the need for an alternate intersection-related data source that provides maximum coverage of intersections while requiring minimal setup cost and maintenance.

Vehicle-based data has recently gained popularity because of its wide coverage, low cost, and minimal maintenance. The network-enabled on-board units (OBUs), which enable vehicles to communicate with external agents in real-time, are a well-known source of vehicle-based data. At the moment, most modern vehicles are equipped with OBUs, allowing them to communicate with their surroundings. Nonetheless, vehicles without OBUs, particularly those that operate in app-based ride-sharing services, can generate vehicle-based data as well. These data-generating vehicles, also known as connected vehicles, as a whole, represent a possible source of appropriate information required for infrastructure-free real-time crash likelihood prediction. Nevertheless, assembling connected vehicle data can be challenging at times. At the industrial level, multiple organizations assemble, and archive connected vehicle data for analysis and visualization. Signal Analytics, primarily operated by INRIX, is one such platform that archives vehicle-based data representing the traffic dynamics at intersections. According to the Regional Integrated Transportation Information System (RITIS), Signal Analytics acquires data from connected vehicles, which account for over 8% of the moving traffic stream. Since connected vehicles are projected to saturate the market soon, Signal Analytics is a viable option for real-time crash likelihood prediction. Regardless of this possibility, to the best of the authors' knowledge, no studies on real-time crash likelihood prediction at intersections so far used vehicle-based data sourced from INRIX and Center for Advanced Transportation Technology (CATT) Lab's Signal Analytics platform for modeling intersections' safety.

From a methodological standpoint, there are two separate ways to predict crash likelihood in real-time: statistical-based and machine learning-based. Although statistical methods are generally focused on identifying possible features that influence crash likelihood, several methods, including logistic regression, conditional logit model, Bayesian random effect logit, etc., have also contributed to real-time crash likelihood prediction. Unfortunately, statistical methods have substantial assumptions and dependencies on data distribution and preparation techniques, which frequently degrade crash likelihood prediction aptitude (Abdel-Aty & Wang, 2006; A. J. M. M. U. Ahmed et al., 2017; Ahsan et al., 2021; F. Guo et al., 2010; Hasan et al., 2022; Lee et al., 2017; Rashid et al., 2018). Machine learning approaches, on the other hand, have strong prediction competencies without the need for any assumptions. Hence, several researchers have applied traditional machine learning methods like support vector machines (Yu & Abdel-Aty, 2013), random forest (Lin et al., 2015; You et al., 2017), etc., to predict crash likelihood. However, the traditional machine learning algorithms frequently fail to handle high-dimensional data (S. Guo et al., 2019). Lately, deep learning algorithms have attracted a lot of attention from traffic safety researchers because of their ability to efficiently work with high-dimensional data. Deep learning is basically a subset of machine learning that focuses on deep neural networks.

Several previous studies have successfully employed deep learning frameworks to predict crash likelihood in real-time. Basso et al. (2021), for example, presented a concatenated convolution neural network (CNN) for predicting real-time crash likelihood in Santiago, Chile freeways. Yu et al. (2020) combined geometry and traffic data gathered from loop detectors on freeways in Shanghai, China, to construct a CNN model with a refined loss function for predicting real-time crash likelihood. Theofilatos et al. (2019) analyzed traffic data from two freeways in Shanghai, China, to compare the predictive performance of multiple machine-learning and deep-learning models, concluding that deep-learning models outperform machine-learning models in real-time crash prediction. To predict real-time crash likelihood on freeways in Florida, Zhang & Abdel-Aty (2022) proposed a bidirectional long short-term memory (LSTM) model with two convolutional layers. Li et al. (2020) developed an LSTM-CNN model for predicting the likelihood of crashes on arterials in Orlando, Florida. Additionally, deep-learning models can be used to augment crash data. For instance, Islam et al. (2021) developed a variational autoencoder to generate crash data for the prediction of crash likelihood. While deep learning models surpassed various traditional machine learning models and statistical models in prediction accuracy in all of the aforementioned studies, the explorations were largely performed on highways and arterials. Deep learning methods have rarely been used to forecast crash likelihood at intersections. To the best of the authors' knowledge, only Yuan et al. (2019) implemented a deep learning technique, the LSTM model, for predicting real-time crash likelihood at signalized intersections. The authors achieved a sensitivity of 60.67%, and a false alarm rate of 39.33%, in favor of the LSTM over the conditional ordered logit model.

In addition to the existing deep neural networks, lately, Transformers have received significant interest among researchers in various fields. The Transformer model, introduced by Vaswani et al. (2017), is an attention-based model, devoid of recurrence and convolutions. Transformers were originally designed for machine translation, but they have since extensively reshaped the field of natural language processing (Devlin et al., 2018; Vaswani et al., 2017) and even extended their reach to other domains such as computer vision (Abdelraouf et al., 2022), speech detection (Gulati et al., 2020). Structurally, Transformer learns dependencies in sequential data by leveraging the attention mechanisms that allow the algorithm to focus directly on any part of the sequence, regardless of the distance. This attribute is, however, missing in sequential algorithms like LSTM. Hence, unlike Transformer, LSTM often fails to handle long-range dependencies in a sequence, resulting in information loss over very long sequences. Another advantage of Transformer over LSTM is, Transformer, avoiding sequential processing like LSTM, can parallelly process all elements in the input sequence during training. Although this parallel processing significantly optimizes the computation time of Transformer, the inability of sequential processing, however, frequently makes it difficult for Transformer to understand dependencies and interactions in temporal sequence data in time series analysis. To circumvent this sequential constraint, researchers have proposed numerous Transformer variants that may effectively execute time series tasks (Xu et al., 2020). Most of the proposed architectures embedded new layers or modified existing layers to precisely capture the temporal order of the sequence in the time series data, resulting in enhanced performance. Unfortunately, despite the immense potential of Transformers in time series analysis, no studies have yet applied Transformer in real-time crash likelihood prediction, which is fundamentally a time series classification task.

In summary, by reviewing existing literatures on real-time crash likelihood prediction three research gaps can be identified. First, to the best of the author's knowledge, connected vehicle data accounting for intersection traffic dynamics have not yet been explored for real-time crash likelihood prediction at intersections. Second, no studies have applied Transformer model to predict crash likelihood in real-time for any roadway segments including intersections. Third, to date there are no viable models that have been developed to predict crash likelihood at intersections in real-time. To fill in these gaps, this study contributed as follow:

- developed time-embedded Transformer model, named intersection-Transformer (inTformer), purposed to predict real-time crash likelihood at intersections.
- explored connected vehicle data sourced from INRIX and CATT Lab's Signal Analytics platform for crash likelihood prediction at intersections.
- identified the contribution of explanatory features on the likelihood of crash occurrence at intersections using SHAP (SHapley Additive exPlanations) approach.

## MODELING CONCEPTION

At intersections, traffic streams from multi-approach directions continuously interact to perform "left," "through," or "right" movements. Hence, unlike roadway segments such as freeways and arterials, the traffic operation mechanism at intersections is comparatively more intricate. This intricacy is further compounded when moving traffic from multi-approach directions fails to comply with the operational imperatives at intersections, demonstrating irrational travel behavior that leads to the potential of traffic crashes. In other words, identifying the relationship between traffic flow patterns and crash occurrence is more challenging at intersections compared to freeways or arterial segments. As a result, in most cases, developing a robust crash likelihood prediction model for intersections is considerably difficult. To address this difficulty, this study proposes a zone-specific modeling approach for crash likelihood prediction at intersections. Under this approach, the intersection area is initially segmented into distinct zones as per comparable traffic flow patterns. Next, traffic crashes in each distinct zone are individually analyzed before finally formulating zone-specific crash likelihood prediction models for intersections.

The remaining sections of this paper extend the modeling idea conceptualized in this section, with the aim of developing real-time crash likelihood prediction models for intersections.

## METHODOLOGY

In this study, to develop the zone-specific real-time crash likelihood model the authors propose inTersection-Transformer (or inTformer), an improved version of the vanilla Transformer adjusted for time-series classification task. The efficacy of the proposed inTformer model to predict crash likelihood at intersections was evaluated against four distinct deep learning approaches: LSTM, CNN, sequential LSTM-CNN, and parallel LSTM-CNN, in terms of several performance metrics.

### Models

#### *inTersection-Transformer / inTformer*

Originally, the Transformer algorithm was designed to perform Natural Language Processing (NLP) tasks by leveraging attention-mechanism (Vaswani et al., 2017). In this study, the functional domain of the original transformer is extended to predict crash likelihood in real-time at intersections. The proposed inTersection-Transformer (inTformer) architecture employed in this paper is depicted in **Figure 1**. The inTformer requisites normalized sequential data in dimension batch_size × sequence_length (timesteps) × input_feature as input. A brief description of each layer in **Figure 1** is summarized below.

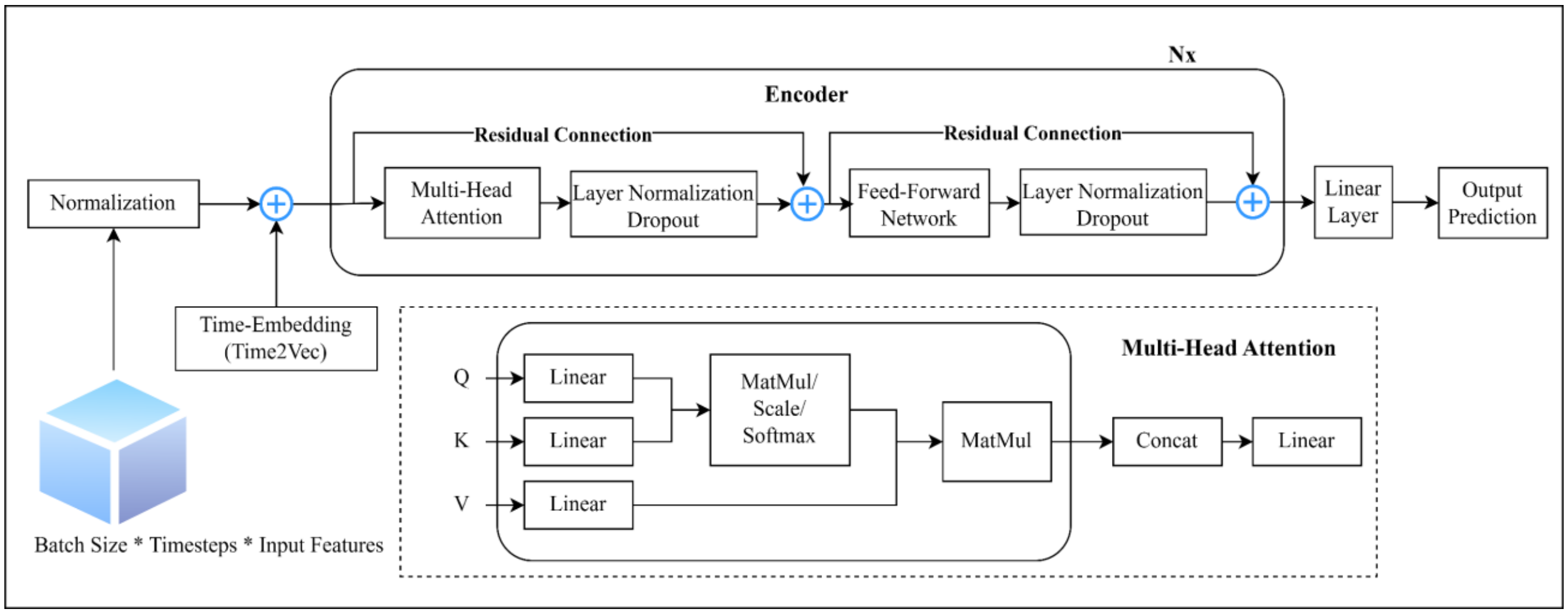

**Figure 1. inTformer Architecture**

*Time Embedding*

In Transformers, all data is forwarded all at once through the model architecture to learn dependencies and interactions in the sequential data. Hence, unlike traditional RNNs and LSTM models, Transformers structurally favor an attention-based system by averting sequential processing. Though the attention-based system can identify dependencies in almost every type of sequential data, at times the inability of sequential processing makes it difficult for Transformer to extract dependencies in data with a time sequence (i.e., time series data). Since real-time crash likelihood prediction requires time series data as input, in this study a serious sequential challenge was posed while developing the inTformer architecture. This sequential issue was resolved by embedding a 'Time Embedding' layer in the inTformer architecture that could account for the temporal order of the data sequence to predict crash within the next 15-30 min.

In the inTformer architecture, the 'Time Embedding' layer was formulated by adopting the model-agnostic time representation, also called 'Time2Vec,' approach proposed by Kazemi et al. (Kazemi et al., 2019). In accordance with the principles proposed in 'Time2Vec,' two ideas were implemented in the 'Time Embedding' layer: *firstly*, a realistic depiction of time must incorporate periodic and nonperiodic patterns, and *secondly*, a time depiction should be invariant to time rescaling, which means that it is unaffected by varied time increments (i.e., seconds, hours, or days) and long-time horizons. The mathematical definition attained by combining the ideas is presented below (**Equation 1**).

$$t2v(\tau)[i] = \begin{cases} \omega_i\tau + \varphi_i, & i = 0 \\ F(\omega_i\tau + \varphi_i), & 1 \leq i \leq k \end{cases} \qquad (1)$$

where $\omega_i\tau + \varphi_i$ represents the non-periodic/linear and $F(\omega_i\tau + \varphi_i)$ the periodic feature of the time vector. $\omega$ in $\omega_i\tau + \varphi_i$ is a matrix that defines the slope of time-series $\tau$ and $\varphi$ in simple terms is a matrix that defines where the time-series $\tau$ intersects with the y-axis. $F(.)$ is a function that makes the linear term $\omega_i\tau + \varphi_i$ periodic.

*Encoder*

The 'Encoder' layer is the core layer of the proposed inTformer. In the seminal paper (Vaswani et al., 2017), the proposed Transformer had two core layers: Encoder and Decoder. However, in this paper, the inTformer focused on leveraging only the functionality of the former to predict crash likelihood at intersections. The proposed inTformer architecture can have multiple 'Encoder' layers, and each 'Encoder' layer incorporates two key sub-layers: Multi-Head Attention Mechanism, and Position-Wise Feed-Forward Network.

Multi-Head Attention Mechanism

This sub-layer executes the attention mechanism of the inTformer by concatenating the attention weights of single heads. Each single-head takes three inputs, namely query $Q$, key $K$, and value $V$, in total to calculate the attention weights that measure the relationship between elements/inputs in a sequence. The $Q$, $K$, and $V$ vectors are obtained by transforming each input from a sequence of inputs (in our case, a sequence of time-embedded inputs).

Say, $X = [x_1, x_2, x_3, \dots, x_n]$ represents time-embedded input sequences, where $x_i$ is the input at timestep $i$. The input $x_i$ can be transformed into three vectors as follows **(Equation 2**):

$$Q_i = x_iW_q; K_i = x_iW_k; V = x_iW_v \qquad (2)$$

where $W_q$, $W_k$, and $W_v$ are weight matrices for the query, key, and value transformations, respectively, which are learned during training. After the $Q$, $K$, and $V$ vectors of all timesteps have been determined, the attention scores are then computed for each pair of $Q$ and $K$ vectors. Specifically, for $Q_i$ and $K_j$ (corresponding to timesteps $i$ and $j$), the score is calculated as the dot product of $Q_i$ and $K_j$ as follows (**Equation 3**):

$$S_{ij} = Q_i K_j^{Transpose} \qquad (3)$$

The attention score $S_{ij}$ measures the similarity between the $Q$ and $K$, effectively determining how much attention should be paid from $Q_i$ (query at time step $i$) to $K_j$ (key at time step $j$). The raw attention scores are then normalized using a 'softmax' function to ensure that they sum up to one across all time steps for each $Q$. Also, the scores are usually scaled down by the square root of the dimension of the key vectors $d_k$ to avoid extremely large values, which could lead to unstable gradients as follows (**Equation 4**):

$$A_{ij} = Softmax\left(\frac{S_{ij}}{\sqrt{d_k}}\right) \qquad (4)$$

For each single-head, using the attention weights, outputs for each input $x_i$ are computed as a weighted sum of the $V$ vectors (**Equation 5**). Finally, the output matrices from all the $h$ single-heads are then concatenated (**Equation 6**) and passed through a final linear transformation (**Equation 7**).

$$O_i = \sum_j A_{ij} \times V_j, \qquad j = 1, 2, \dots, n \text{ including timestep } i \qquad (5)$$
$$O_{concat} = concat(O_1, O_2, \dots, O_h) \qquad (6)$$
$$O_{final} = linear(O_{concat}) \qquad (7)$$

The final output vectors $O_{final}$ form the output sequence of the multi-head attention mechanism.

Position-Wise Feed-Forward Network

The purpose of this sub-layer is to transform the representation received from the 'Multi-Head Attention Mechanism,' allowing the model to identify more complex relationships. The term "position-wise" in 'Position-Wise Feed-Forward Network' refers to the fact that the same Feed-Forward Network is applied separately to each input in a sequence of inputs (in our case, a sequence of time-embedded inputs). This is analogous to using a convolution with a kernel size of 1 in CNN. As a result, CNN with convolution kernel size 1 was incorporated into the inTformer architecture as a 'Position-Wise Feed-Forward Network.' The CNN used in inTformer's 'Encoder' includes two layers, which are as follows:

1. In the *first* layer, initially a simple linear transformation takes place that projects (increases) the dimension of input data to a higher dimensional space. Following the linear transformation, an element-by-element application of a non-linear activation function (in our example, ReLU) is performed.
2. In the *second* layer, the output from the activation function is then passed through a second linear transformation, which projects the data back to the original dimension.

In 'Encoder,' each of the key sub-layers: 'Multi-Head Attention Mechanism' and 'Position-Wise Feed-Forward Network,' are followed by dropout, residual connection, and layer normalization. The inclusion of dropout helps prevent the model from overfitting by not allowing it to rely too heavily on any single input/element in the sequence. The residual connections, also known as skip or shortcut connections, help the inTformer combat the problem of vanishing gradients. Layer normalization is a process that normalizes the values of the activations in a layer to have a mean of 0 and a variance of 1. This helps keep the activations and gradients on a similar scale, leading to more stable training. In inTformer, layer normalization helps to ensure that the scale of the values throughout the model doesn't get out of control, which can lead to issues with learning.

***LSTM***

The LSTM architecture (Hochreiter & Schmidhuber, 1997) is a recurrent neural network (RNN) architecture designed to model sequential input. Previously, traditional RNNs experienced vanishing gradient problems when learning large data sequences (Arbel, 2018). LSTM, by incorporating the memory cell to determine when to forget certain information, could solve this problem.

The LSTM network employs a series of connected cells. An input gate $i_t$, a forget gate $f_t$, an output gate $o_t$, a memory cell $c_t$, and a hidden state $h_t$, comprises a standard LSTM cell. The input and forget gates determine how much information is drawn from the current timestep features $x_t$ and how much information is drawn from the preceding cell hidden state $h_{t-1}$. The output gate determines how much information is transferred to the next cell by calculating the current cell's hidden state $h_t$. The calculations performed by each LSTM cell at timestep $t$ are specified in **Equations 8-13** (Graves et al., 2013). The computations are carried out for each member of the modeled sequence $t = 0...T$.

$$i_t = \sigma(W_{ix}X_t + W_{ih}h_{t-1} + W_{ic}c_{t-1} + b_i) \quad (8)$$
$$f_t = \sigma\left(W_{fx}X_t + W_{fh}h_{t-1} + W_{fc}c_{t-1} + b_f\right) \quad (9)$$
$$o_t = \sigma(W_{ox}X_t + W_{oh}h_{t-1} + W_{oc}c_{t-1} + b_o) \quad (10)$$
$$c_t = f_t \odot c_{t-1} + i_t \odot tanh(W_{cx}X_t + W_{ch}h_{t-1} + b_c) \quad (11)$$
$$h_t = o_t \odot tanh(c_t) \quad (12)$$
$$y_t = W_{yh}h_{t-1} + b_y \quad (13)$$

where $W$, and $b$ represents network trainable weight matrices, and bias vectors, respectively. $\sigma$ is the logistic sigmoid function, and $\odot$ indicates the elementwise product of the vectors.

***CNN***

CNN was first designed to address image classification challenges. Recently, multiple studies suggested that CNN may also be leveraged to learn time-series data with promising results (Wang et al., 2016; Zhao et al., 2017). The convolution layer of CNN employs a filter to extract features from the input data (Fawaz et al., 2019). As a feature extractor, this work used a one-dimensional (1D) CNN using Rectified Linear Unit (ReLU) as the activation function. The capacity of CNN to learn features that are invariant across time dimensions is one advantage of employing it for time-series data. Moreover, previous studies (Li et al., 2020; Li & Abdel-Aty, 2022) also validated the possibility of utilizing CNN for crash likelihood prediction.

## Performance Metrics

Generally, several key metrics, such as accuracy, sensitivity, false positive rate, etc., are utilized to evaluate predictive performance of deep neural networks. While accuracy is often assessed for balanced datasets, model performance for imbalanced datasets is frequently measured using metrics such as sensitivity, false alarm rate, and so on. Hence, in this study, sensitivity, and false alarm rate were employed for model performance evaluation using test dataset. Sensitivity (**Equation 14**) and false alarm rate (**Equation 15**) are derived from the classification confusion matrix (**Table 1**). The sensitivity reflects the percentage of correct positive predictions produced (Raihan et al., 2023; Syed et al., 2023), whereas the false positive rate indicates how frequently the model is likely to mispredict.

**Table 1. Confusion Matrix**

| | | Predicted Label | |
|---|---|---|---|
| | | **Crash** | **Non-Crash** |
| **Actual Label** | **Crash** | True Positive (TP) | False Negative (FN) |
| | **Non-Crash** | False Positive (FP) | True Negative (TN) |

$$True\ Positive\ Rate\ (Sensitivity) = \frac{TP}{TP + FN} \quad (14)$$

$$False\ Positive\ Rate\ (False\ Alarm\ Rate) = \frac{FP}{FP + TN} \quad (15)$$

## Model Interpretation

Machine learning models, while adept at providing precise predictions, are frequently labeled as "black boxes," making it challenging to interpret the impact of explanatory features on predictions. In response to this challenge, Lundberg and Lee (Lundberg & Lee, 2017) introduced a game-theoretic strategy known as SHAP (SHapley Additive exPlanations). This approach offers a way to measure the importance of features and analyze their effects on the predictions in machine learning architectures, including deep neural networks. In this study, the SHAP approach was applied to identify the importance and contribution of features within the inTformer models. For a given feature, denoted as $i$, its influence and importance, as depicted by the SHAP value, are derived as follows (**Equation 16**):

$$\varphi_i = \sum_{S \subseteq F - i} \frac{|S|!\,(|F| - |S| - 1)!}{|F|!} \left[ f_{S \cup \{i\}}\left(x_{S \cup \{i\}}\right) - f_S(x_S) \right] \quad (16)$$

where $|F|$ is the total number of explanatory features, $S$ represents any subset of explanatory features that doesn't include the $i^{th}$ feature and $|S|$ is the size of that subset. $f_{S \cup \{i\}}\left(x_{S \cup \{i\}}\right)$ indicates model trained with $i$, and $f_S(x_S)$ is model trained without $i$.

# DATA DESCRIPTION AND PREPARATION

## Data Description

In general, the frequency of crashes at a location is a direct reflection of the riskiness of the location, i.e., the more frequently crashes occur, the more dangerous the location. Given this fact, this paper selected six hazardous intersections in Tampa and two in Orlando, Florida, as shown in **Table 2**. For the period July 2021 to June 2022, three types of datasets were collected in this study: (a) crash data from Signal Four Analytics (S4A); (b) connected vehicle data from INRIX and CATT Lab's Signal Analytics; (c) weather data from Visual Crossings.

**Table 2. Selected Intersections with Crash Count**

| Intersection Name | Latitude | Longitude | Crash Count |
|---|---|---|---|
| East Hillsborough Avenue | 27.99616 | -82.45416 | 75 |
| West Brandon Boulevard & Brandon Town Center Drive | 27.93939 | -82.32389 | 66 |
| East Dr. Martin Luther King Jr Boulevard & North Marguerite Street | 27.98149 | -82.45422 | 60 |
| Polk City Road & US 27 | 28.12132 | -81.6397 | 57 |
| East Hillsborough Avenue & North Nebraska Avenue | 27.99607 | -82.45114 | 54 |
| Glen Este Boulevard & US 27 | 28.12499 | -81.6397 | 51 |
| East Dr. Martin Luther King Jr Boulevard & US 301 | 27.98146 | -82.36012 | 50 |
| West Columbus Drive & North Dale Mabry Highway | 27.96679 | -82.50551 | 49 |

***Crash Data***

This paper employed crash data retrieved from the Signal Four Analytics (S4A) database. The crash data from S4A includes information on the crash timing, type, severity, and location, as well as other features such as weather, road conditions, etc. In this research, all crashes that occurred at the study intersections or road sections influenced by intersections (within 250 feet (76.2 meter) of intersections (Yuan et al., 2019; Yuan & Abdel-Aty, 2018)) between July 2021 and June 2022 were collected. Using QGIS, the collected crashes were then matched with the respective intersections. In total, 462 crashes were jointly identified within 250 feet (76.2 meter) from all study intersections.

***Connected Vehicle Data***

In this study, the traffic data were collected from the Signal Analytics platform. Signal Analytics is the product of a collaboration between INRIX and the Center for Advanced Transportation Technology (CATT) Lab that focuses on capturing the traffic dynamics at intersections. Signal Analytics collects real-time traffic data from connected vehicles. To do so, each sampled vehicle is assigned to one of the 230-meter zones (one for each approach direction), as shown in **Figure 2**. Once a connected vehicle enters the 150-meter 'Shared Approach Zone' in **Figure 2**, Signal Analytics begins collecting pings from the vehicle every 3 to 5 seconds until the vehicle exits via the 80-meter 'Right,' 'Through,' or 'Left,' zone. Following ping collection, the Signal Analytics platform estimates travel times, control delays, and whether a vehicle is considered to have stopped in the intersection analysis zone. In total, the platform stores aggregated information on nine different features, namely, split failure count (SFC) and split failure percentage (SFP), travel time maximum (TTM) and travel time average (TTA), control delay maximum (CDM) and control delay average (CDA), approach speed maximum (ASM) and approach speed average (ASA), and percent (arrival) on green (POG) every 15 min.

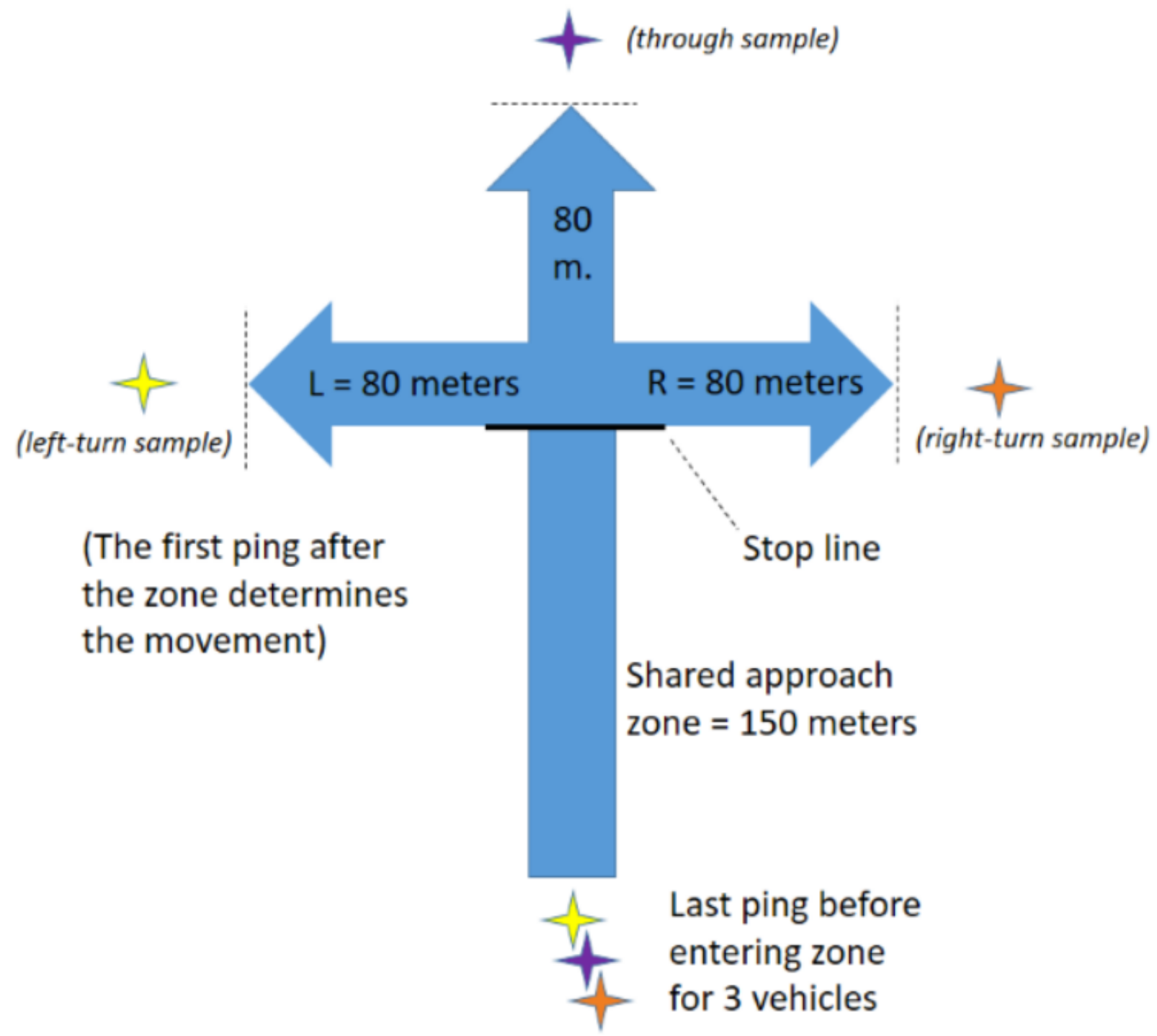

**Figure 2. Intersection Analysis Zone** (*Signal Analytics Help*)

A split failure happens when the green signal time fails to meet the vehicle volume demand during one cycle, i.e., if a driver has to wait for more than one cycle at a traffic light. The trip time is based on the time it takes each observed vehicle to traverse the intersection's 230-meter analysis zone. The control delay is the difference between the observed and reference travel times; the reference travel time is the $5^{th}$-percentile fastest travel time recorded for each intersection approach direction. The approach speed is the calculated speed of each sampled vehicle while traversing the 150-meter approach zone. Finally, the percentage of observed vehicles that passed through the crossing without stopping is designated as the POG. POG can be expressed mathematically as follows (**Equation 17**):

$$POG = \frac{Vehicle\ Count: Total - Vehicle\ Count: Stopped}{Vehicle\ Count: Total} \times 100 \qquad (17)$$

***Weather Data***

Six weather-related features (temperature, relative humidity, wind speed, precipitation, visibility, and weather type) were retrieved using weather API from Visual Crossings. The API provides GPS location-specific weather information by triangulating data from nearby weather stations as well as weather radars.

## Data Preparation

***Intersection Zoning***

Typically, the way traffic behaves at the approaches to intersections differs from the way traffic behaves inside the intersections. Inside the intersection region, where the approach roads physically intersect, traffic interacts more directly i.e., traffic of all approach roads simultaneously act to make turns, or cross paths. However, in the approach region, which only includes the sections of road that lead up to the intersection, incoming traffic rarely execute turning or crossing movements, thus avoiding direct interactions. Acknowledging this regional disparity in traffic flow patterns, this study divided intersection area into two zones: within-intersection, and approach, to facilitate the proposed zone-specific crash modeling technique. The zone-specific crash prediction models were developed by analyzing crashes in the respective 'within-intersection' and 'approach' zone.

**Figure 3** depicts the boundary of the distinct intersection zones. The 'within-intersection' zone is simply the rectangular zone at the intersection that connects all of the approaches (represented by the 'red' rectangle). To determine the boundary of the 'approach' zone, prior studies were followed (Yuan et al., 2019; Yuan & Abdel-Aty, 2018). The existing studies implied that intersections have a radial influence of 250 feet (76.2 meter) from their center to its surroundings. Hence, in this study, the approach road segments falling within the intersection's influence margin, i.e., within 250 feet (76.2 meter), were designated as the 'approach' zone (represented by 'blue' rectangles).

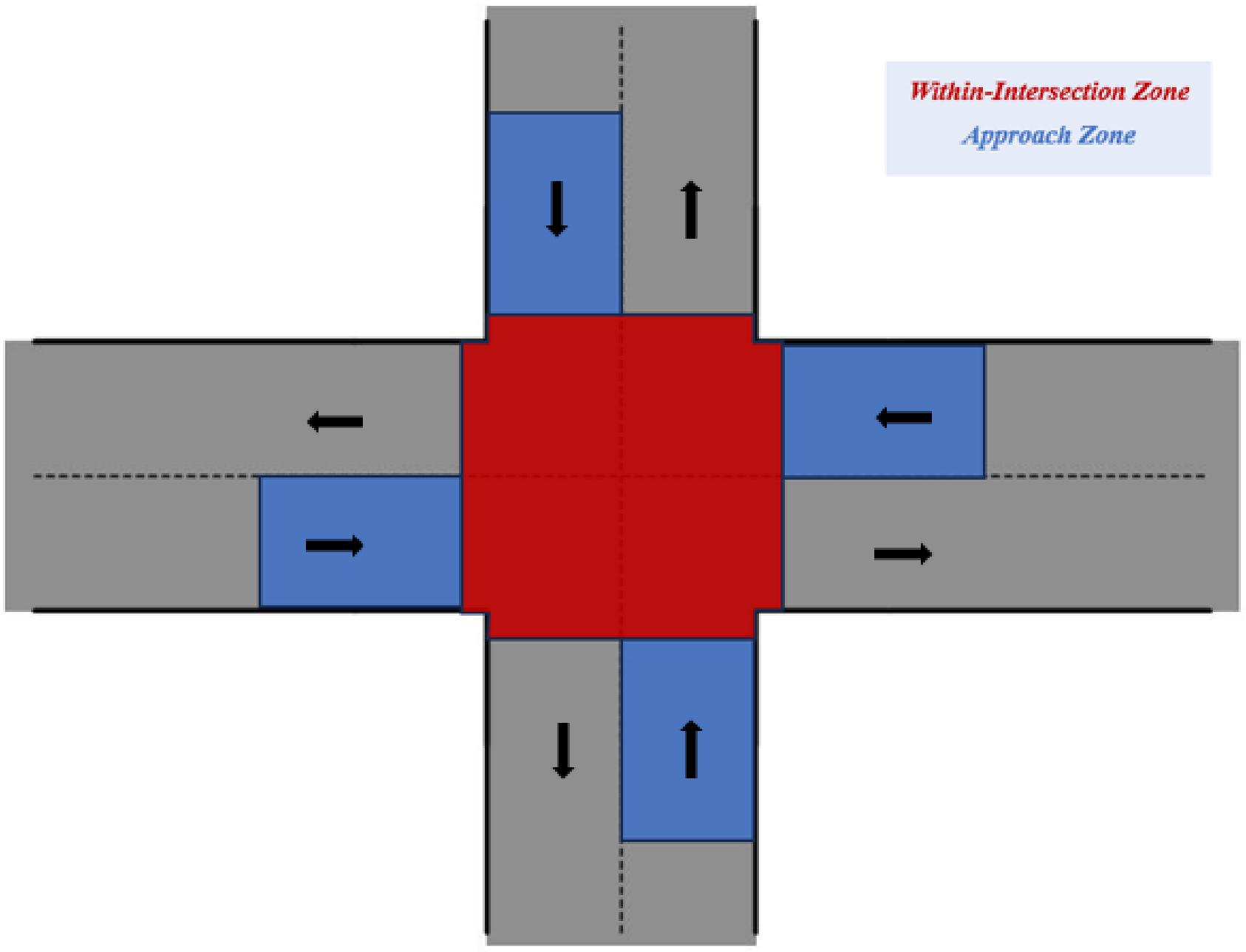

**Figure 3. Intersection Zoning**

***Traffic Data Formatting***

In this study, real-time traffic data from the Signal Analytics archive, which is updated every 15 minutes, were retrieved for all intersections. Using the retrieved raw traffic data, a total of 1,051,200 observations (30 ((6 four-legged intersection × 4 approaches) + (2 three-legged intersections × 3 approaches)) approaches × 365 days × 24 hours × (60 minutes/15 minutes)) were generated, where each observation corresponds to the traffic features of the approach under consideration at a given intersection over a span of 15 minutes. Next, the weather data was matched and integrated with each generated observation by taking intersection location and time into account. The combined traffic and weather data were then formatted for 'within-intersection' and 'approach' zones as follows:

As stated earlier, in the 'within-intersection' zone, traffic of all approaches to any given intersection interact more directly to accomplish turning or crossing movements. As a result, in most cases, the traffic conditions of all approach directions directly or indirectly influence the mechanisms leading to crash occurrences in the 'within-intersection' zone. Hence, to precisely capture the relationship between traffic flow patterns and crashes at the 'within-intersection' zone, it is essential to structure the dataset in a way that guarantees each 15-minute observation period includes traffic features of all approaches leading to an intersection. To formulate such dataset, this study adopted the direction nomenclature proposed by Yuan & Abdel-Aty (2018). In accordance with the nomenclature, the intersection approaches were renamed as 'A,' 'B,' 'C,' and 'D,' where 'A' approach denotes the approach under consideration. The 'B' approach denotes the 'A' approach's left-side approach, whereas the 'C,' and 'D' approaches follow a clockwise naming, respectively (see **Figure 4**). Upon formatting, the initially generated dataset of 1,051,200 observations didn't only have the traffic features of the approach under consideration at a given intersection, plus the weather features. Instead, the dataset incorporated the traffic features of all approaches leading to the given intersection and the weather features. **Table 3** presents the summary statistics of traffic and weather features used in the 'within-intersection' crash analysis. It is to note that, for illustrative purposes, only the traffic features of approach 'A' are presented in **Table 3**. In the final crash analysis, traffic features of all approaches ('A,' 'B,' 'C,' and 'D') were used considered.

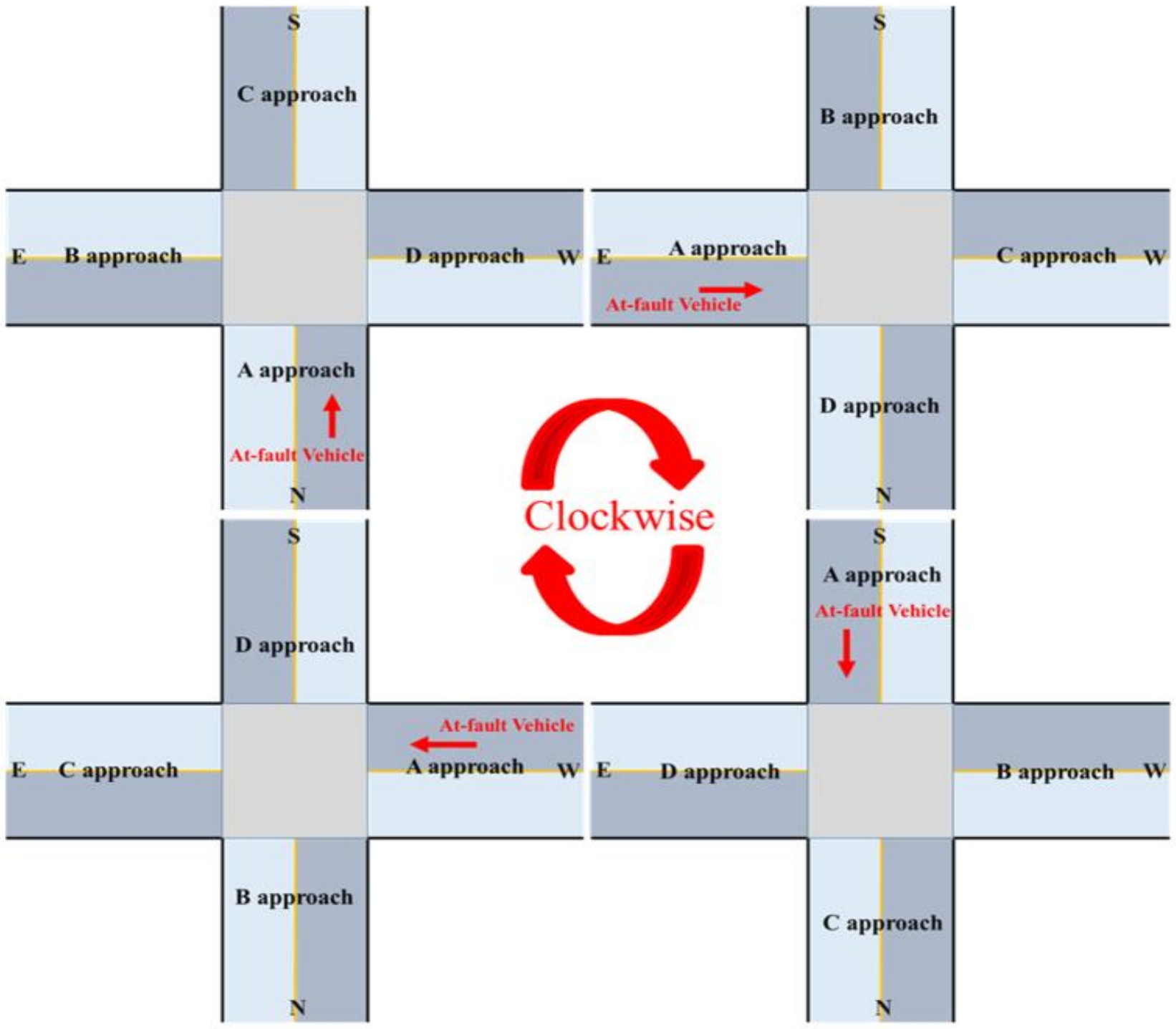

**Figure 4. Direction Nomenclature** (Yuan & Abdel-Aty, 2018)

In contrast to the 'within-intersection' zone, the crash occurrence mechanism in the 'approach' zone solely relies on the traffic flow patterns of the approach where crashes take place. To put it more precisely, crashes happening at a given approach to an intersection are only influenced by the traffic conditions of that particular approach, not by the traffic conditions of the rest of the approaches leading to the given intersection. Given this fact, the dataset necessary for

**Table 3. Feature Descriptive Statistics of 'Within-Intersection' Zone**

| Feature | Description | Unit | Mean(Std) | (Min, Max) |
|---|---|---|---|---|
| *ASA_L_A* | Average Speed of Left-Turning, Through, and Right-Turning Vehicles at Approach 'A' | mph | 30.13(5.43) | (11.0, 77.0) |
| *ASA_T_A* | | | 34.87(7.2) | (9.0, 103.0) |
| *ASA_R_A* | | | 31.89(5.29) | (6.0, 102.0) |
| *ASM_L_A* | Maximum Speed of Left-Turning, Through, and Right-Turning Vehicles at Approach 'A' | mph | 31.39(5.68) | (11.0, 78.0) |
| *ASM_T_A* | | | 42.26(11.85) | (9.0, 147.0) |
| *ASM_R_A* | | | 33.34(5.51) | (6.0, 100.0) |
| *TTA_L_A* | Average Travel Time of Left-Turning, Through, and Right-Turning Vehicles at Approach 'A' | second | 71.53(45.85) | (8.0, 532.0) |
| *TTA_T_A* | | | 45.32(34.32) | (5.0, 500.0) |
| *TTA_R_A* | | | 32.71(21.8) | (7.0, 413.0) |
| *TTM_L_A* | Maximum Travel Time of Left-Turning, Through, and Right-Turning Vehicles at Approach 'A' | second | 82.16(55.24) | (8.0, 556.0) |
| *TTM_T_A* | | | 68.59(43.11) | (5.0, 570.0) |
| *TTM_R_A* | | | 37.52(27.9) | (7.0, 489.0) |
| *CDA_L_A* | Average Control Delay of Left-Turning, Through, and Right-Turning Vehicles at Approach 'A' | second | 57.94(45.0) | (1.0, 378.0) |
| *CDA_T_A* | | | 34.58(32.39) | (0.0, 491.0) |
| *CDA_R_A* | | | 17.37(21.83) | (0.0, 395.0) |
| *CDM_L_A* | Maximum Control Delay of Left-Turning, Through, and Right-Turning Vehicles at Approach 'A' | second | 68.4(54.3) | (1.0, 543.0) |
| *CDM_T_A* | | | 57.83(43.04) | (1.0, 559.0) |
| *CDM_R_A* | | | 22.27(28.01) | (1.0, 473.0) |
| *SFC_L_A* | Count of Split Failure of Left-Turning, Through, and Right-Turning Vehicles at Approach 'A' | - | 0.09(0.37) | (0, 8) |
| *SFC_T_A* | | | 0.02(0.19) | (0, 9) |
| *SFC_T_A* | | | 0.04(0.23) | (0, 7) |
| *SFP_L_A* | Percentage of Split Failure of Left-Turning, Through, and Right-Turning Vehicles at Approach 'A' | % | 3(12) | (0, 100) |
| *SFP_T_A* | | | 1(5) | (0, 100) |
| *SFP_R_A* | | | 1(6) | (0, 100) |
| *POG_L_A* | Percentage of Left-Turning, Through, and Right-Turning Vehicles Arrived on Green at Approach 'A' | % | 22(32) | (0, 100) |
| *POG_T_A* | | | 54(36) | (0, 100) |
| *POG_R_A* | | | 69(34) | (0, 100) |
| *Temperature* | Dry-bulb Temperature | Fahrenheit | 74.82(11.19) | (41.5, 95.0) |
| *Relative Humidity* | Relative Humidity | % | 68.51(17.56) | (20.6, 100) |
| *Wind Speed* | Speed of the Wind | mph | 5.89(3.73) | (0.0, 19.9) |
| *Precipitation* | Amount of Precipitation | inches to hundredths | 0.0(0.04) | (0.0, 0.72) |
| *Visibility* | Horizontal Distance an Object can be Seen | miles | 9.56(1.11) | (0.6, 9.9) |
| *Conditions* | Normal Weather: 0, Abnormal Weather: 1 | - | 0.18(0.38) | (0, 1) |

*Here, A: Approach "A," L: Left-Turn, T: Through, and R: Right-Turn.*
*Note:* ***Table 3 only depicts Approach "A" traffic data for illustrative reasons. In the final crash analysis, traffic features of all approaches were considered.***

**Table 4. Feature Descriptive Statistics of 'Approach' Zone**

| Feature | Description | Unit | Mean(Std) | (Min, Max) |
|---|---|---|---|---|
| *ASA_L* | Average Speed of Left-Turning, Through, and Right-Turning Vehicles | mph | 30.13(5.43) | (11.0, 77.0) |
| *ASA_T* | | | 34.87(7.2) | (9.0, 103.0) |
| *ASA_R* | | | 31.89(5.29) | (6.0, 102.0) |
| *ASM_L* | Maximum Speed of Left-Turning, Through, and Right-Turning Vehicles | mph | 31.39(5.68) | (11.0, 78.0) |
| *ASM_T* | | | 42.26(11.85) | (9.0, 147.0) |
| *ASM_R* | | | 33.34(5.51) | (6.0, 100.0) |
| *TTA_L* | Average Travel Time of Left-Turning, Through, and Right-Turning Vehicles | second | 71.53(45.85) | (8.0, 532.0) |
| *TTA_T* | | | 45.32(34.32) | (5.0, 500.0) |
| *TTA_R* | | | 32.71(21.8) | (7.0, 413.0) |
| *TTM_L* | Maximum Travel Time of Left-Turning, Through, and Right-Turning Vehicles | second | 82.16(55.24) | (8.0, 556.0) |
| *TTM_T* | | | 68.59(43.11) | (5.0, 570.0) |
| *TTM_R* | | | 37.52(27.9) | (7.0, 489.0) |
| *CDA_L* | Average Control Delay of Left-Turning, Through, and Right-Turning Vehicles | second | 57.94(45.0) | (1.0, 378.0) |
| *CDA_T* | | | 34.58(32.39) | (0.0, 491.0) |
| *CDA_R* | | | 17.37(21.83) | (0.0, 395.0) |
| *CDM_L* | Maximum Control Delay of Left-Turning, Through, and Right-Turning Vehicles | second | 68.4(54.3) | (1.0, 543.0) |
| *CDM_T* | | | 57.83(43.04) | (1.0, 559.0) |
| *CDM_R* | | | 22.27(28.01) | (1.0, 473.0) |
| *SFC_L* | Count of Split Failure of Left-Turning, Through, and Right-Turning Vehicles | - | 0.09(0.37) | (0, 8) |
| *SFC_T* | | | 0.02(0.19) | (0, 9) |
| *SFC_T* | | | 0.04(0.23) | (0, 7) |
| *SFP_L* | Percentage of Split Failure of Left-Turning, Through, and Right-Turning Vehicles | % | 3(12) | (0, 100) |
| *SFP_T* | | | 1(5) | (0, 100) |
| *SFP_R* | | | 1(6) | (0, 100) |
| *POG_L* | Percentage of Left-Turning, Through, and Right-Turning Vehicles Arrived on Green | % | 22(32) | (0, 100) |
| *POG_T* | | | 54(36) | (0, 100) |
| *POG_R* | | | 69(34) | (0, 100) |
| *Temperature* | Dry-bulb Temperature | Fahrenheit | 74.82(11.19) | (41.5, 95.0) |
| *Relative Humidity* | Relative Humidity | % | 68.51(17.56) | (20.6, 100) |
| *Wind Speed* | Speed of the Wind | mph | 5.89(3.73) | (0.0, 19.9) |
| *Precipitation* | Amount of Precipitation | inches to hundredths | 0.0(0.04) | (0.0, 0.72) |
| *Visibility* | Horizontal Distance an Object can be Seen | miles | 9.56(1.11) | (0.6, 9.9) |
| *Conditions* | Normal Weather: 0, Abnormal Weather: 1 | - | 0.18(0.38) | (0, 1) |

*Here, L: Left-Turn, T: Through, and R: Right-Turn.*

predicting crash likelihood in the 'approach' zone only requires the 15-minute observation period to contain traffic features specific to the approach under consideration, plus the weather features. This configuration precisely corresponds with the structure of the initial dataset generated at the start of the traffic data formatting process. As a result, for modeling crashes in the 'approach' zone, the initially formatted data was used. The summary statistics of the traffic and the weather features, for 'approach' zone crash analysis, is depicted in **Table 4**.

In summary, for analyzing crashes in the 'within-intersection' zone, the traffic data was structured to capture traffic-related features from every approach leading to a given intersection. Conversely, for investigating crashes in the 'approach' zone, only the traffic features of the approach under consideration were kept. Nonetheless, in both analyses, the weather-related features remained unchanged.

***Crash Identification***

The S4A crash data archive offers the location of each crash incidence using latitude and longitude coordinates. With this geographic information, the extracted 462 crash incidences were sorted into designated intersection zones. In total, 338 (73.2%) crashes happened in the 'within-intersection' zone, while 124 (26.8%) crashes happened in the approach zone.

***Crash Indexing***

In the process of preparing crash data for real-time modeling, a key step involves assigning crash-likelihood index to each timestep, usually at intervals of 5, 10, or 15 minutes, that signifies whether a crash event has happened or not. This study relied on crash data from S4A to perform such indexing at every 15-minute timestep. To illustrate, if a crash was reported at 10:50 in the crash data, the preceding 15-minute interval at 10:45 was identified as a high-risk period and given an index of 1, representing a crash event. This indicates that a crash is expected within the following 15 minutes (see **Figure 5**). Additionally, this study assumed that the state of traffic 15-30 minutes before the crash could influence the occurrence of the crash as well (Cheng et al., 2022; Yuan et al., 2018). Therefore, in the instance of a crash reported at 10:50, the 10:30 timestep was also given an index of 1. All other timesteps that did not correspond to an actual crash incidence were indexed as 0, denoting a non-crash event.

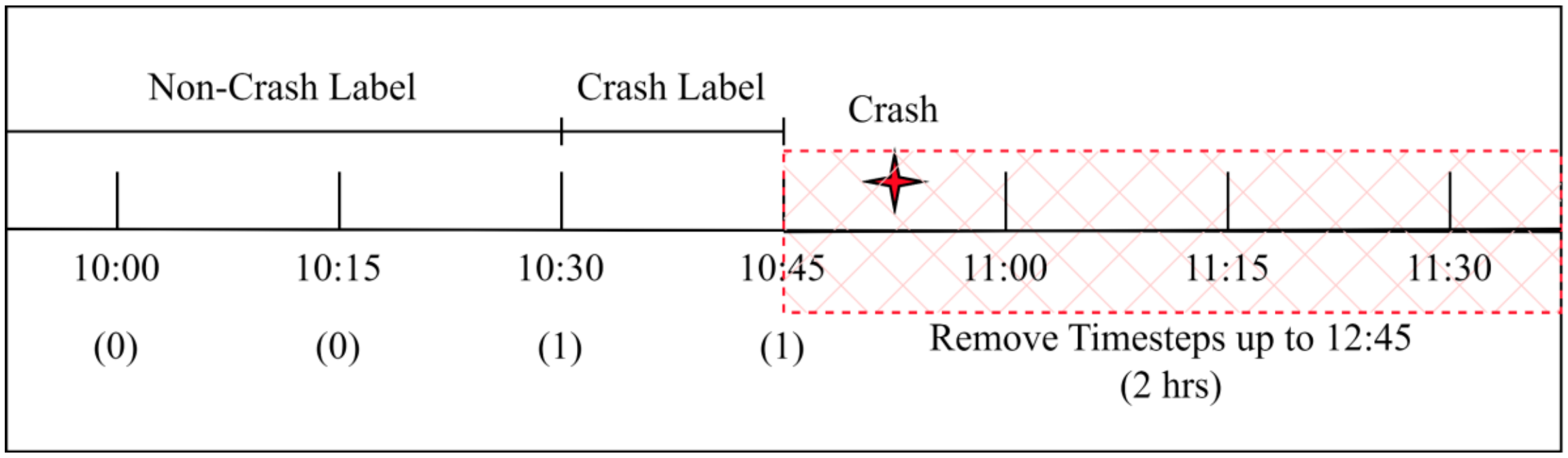

**Figure 5. Crash Indexing**

The indexing was carried out distinctly for crashes in the 'within-intersection' and 'approach' zones. After indexing, the indexed crash data of the 'within-intersection' zone was combined with the formatted traffic dataset of the 'within-intersection' zone. Similarly, the indexed crash data of the 'approach' zone was combined with the formatted traffic dataset of the 'approach' zone. Further, for each defined zone, data (encompassing traffic, weather, and crash information) for two hours following the timestep of the indexed crash event was disregarded in this study. This decision stemmed from existing studies that implied that the occurrence of crash can induce instability in traffic conditions (Li et al., 2022; Li & Abdel-Aty, 2022).

### *Feature Selection*

In this study, extra-tree (Geurts et al., 2006), and Pearson correlation coefficient (Benesty et al., 2009) were employed to assess feature importance and correlation. Using feature importance and correlation coefficient scores, a feature selection rule was devised. As per the rule, if two features have a correlation greater than 0.5, then the most important feature based on the feature importance score was kept for modeling. The feature selection rule was independently applied on data of the 'within-intersection' and 'approach' zones.

Before applying feature selection, the datasets for the 'within-intersection' and 'approach' zones had 114 and 33 independent features (traffic features + weather features), respectively. After the feature selection rule was applied to datasets of both zones, the feature count for the 'within-intersection' zone reduced to 71, while for the 'approach' zone, it came down to 18. The selected feature for each defined zone is presented in **Table 5**.

**Table 5. List of Selected Features**

| Zones | Selected Features |
|---|---|
| Within-Intersection | *'ASA_L_A', 'ASA_T_D', 'ASM_T_D', 'ASM_L_A', 'CDA_L_A', 'TTM_L_A', 'CDM_L_A', 'SFP_L_A', 'ASA_T_A', 'CDA_T_A', 'TTA_T_A', 'TTM_T_A', 'POG_T_A', 'SFP_T_A', 'ASA_R_A', 'ASA_L_B', 'ASM_L_B', 'ASA_T_B', 'ASM_R_A', 'CDA_R_A', 'CDM_R_A', 'TTA_R_A', 'SFP_R_A', 'CDA_L_B', 'TTA_L_B', 'TTM_L_B', 'SFP_L_B', 'CDA_T_B', 'TTA_T_B', 'CDM_T_B', 'POG_T_D', 'SFP_T_B', 'ASM_R_B', 'ASA_R_B', 'ASA_L_C', 'ASM_L_C', 'CDA_R_B', 'TTA_R_B', 'TTM_R_B', 'SFP_R_B', 'CDM_L_C', 'TTA_L_C', 'CDA_L_C', 'SFP_L_C', 'ASA_T_C', 'CDM_T_C', 'TTA_T_C', 'TTM_T_C', 'SFP_T_C', 'ASA_R_C', 'ASM_R_C', 'ASM_L_D', 'CDA_R_C', 'TTA_R_C', 'TTM_R_C', 'ASA_L_D', 'SFP_R_C', 'CDA_L_D', 'TTA_L_D', 'TTM_L_D', 'SFP_L_D', 'CDM_T_D', 'TTA_T_D', 'TTM_T_D', 'SFP_T_D', 'ASA_R_D', 'ASM_R_D', 'CDA_R_D', 'TTA_R_D', 'TTM_R_D', 'SFP_R_D'* |
| Approach | *'ASA_L', 'CDM_L', 'TTA_L', 'TTM_L', 'SFP_L', 'ASM_T', 'CDA_T', 'TTA_T', 'CDM_T', 'TTM_T', 'SFP_T', 'ASM_R', 'ASA_R', 'CDA_R', 'TTA_R', 'TTM_R', 'SFP_R', Precipitation* |

*Here, A: Approach 'A,' B: Approach 'B,' C: Approach 'C,' D: Approach 'D,' and L: Left-Turn, T: Through, R: Right-Turn.*

### *Data Mapping*

The proposed inTformer, like most established deep neural networks, take three-dimensional dataset (batch_size × sequence_length (timesteps) × input_feature) as input. Hence, prior to performing the analysis, the two-dimensional datasets (with traffic, weather, and crash information) for 'within-intersection' and 'approach' zones were independently mapped to three-dimensional datasets through data stacking. The datasets were stacked on timesteps, since previous studies (Li et al., 2022; Yuan et al., 2019) showed that combining 2-5 timesteps prior to crash occurrence increases the chance of accurate crash likelihood prediction. In this study, three combinations of timesteps stacking were tested, though previous studies mostly exercised a single combination. The combinations include stacking *two (batch_size × 2 × input_feature), three (batch_size × 3 × input_feature),* and *four (batch_size × 4 × input_feature),* timesteps in data, where each timestep represents an observation in the two-dimensional dataset, and the stacked timesteps represent an observation in the three-dimensional dataset. This process generated three stacked datasets for each intersection zone.

***Data Sampling***

In relation to non-crash events, the instances of crash events in the mapped datasets for 'within-intersection' and 'approach' zones were extremely low, i.e., the data were highly imbalanced. To address data imbalance, typically, majority or minority classes are often resampled, or the cost function is altered to make the misclassification of minority classes more important than the misclassification of majority classes. In this paper, the resampling technique, particularly the synthetic minority over-sampling technique (SMOTE) (Chawla et al., 2002), was employed to overcome the imbalance problem (Li et al., 2020; Yuan et al., 2019). SMOTE is an over-sampling method that generates new and synthetic data using the nearest neighbor algorithm. It creates new minority instances from existing minority instances. In this study, after splitting the mapped datasets for 'within-intersection' and 'approach' zones into training (75%) and testing (25%) sets, SMOTE was applied on the training sets to generate synthetic samples of crash events in a 1:1 ratio to balance crash and non-crash events. All models including the proposed inTformer, LSTM, CNN, sequential LSTM-CNN, and parallel LSTM-CNN, were trained on synthetic train datasets. To evaluate all the trained crash likelihood prediction models, unsampled test datasets were employed.

The overall data preparation pipeline of this study is depicted in **Figure 6**.

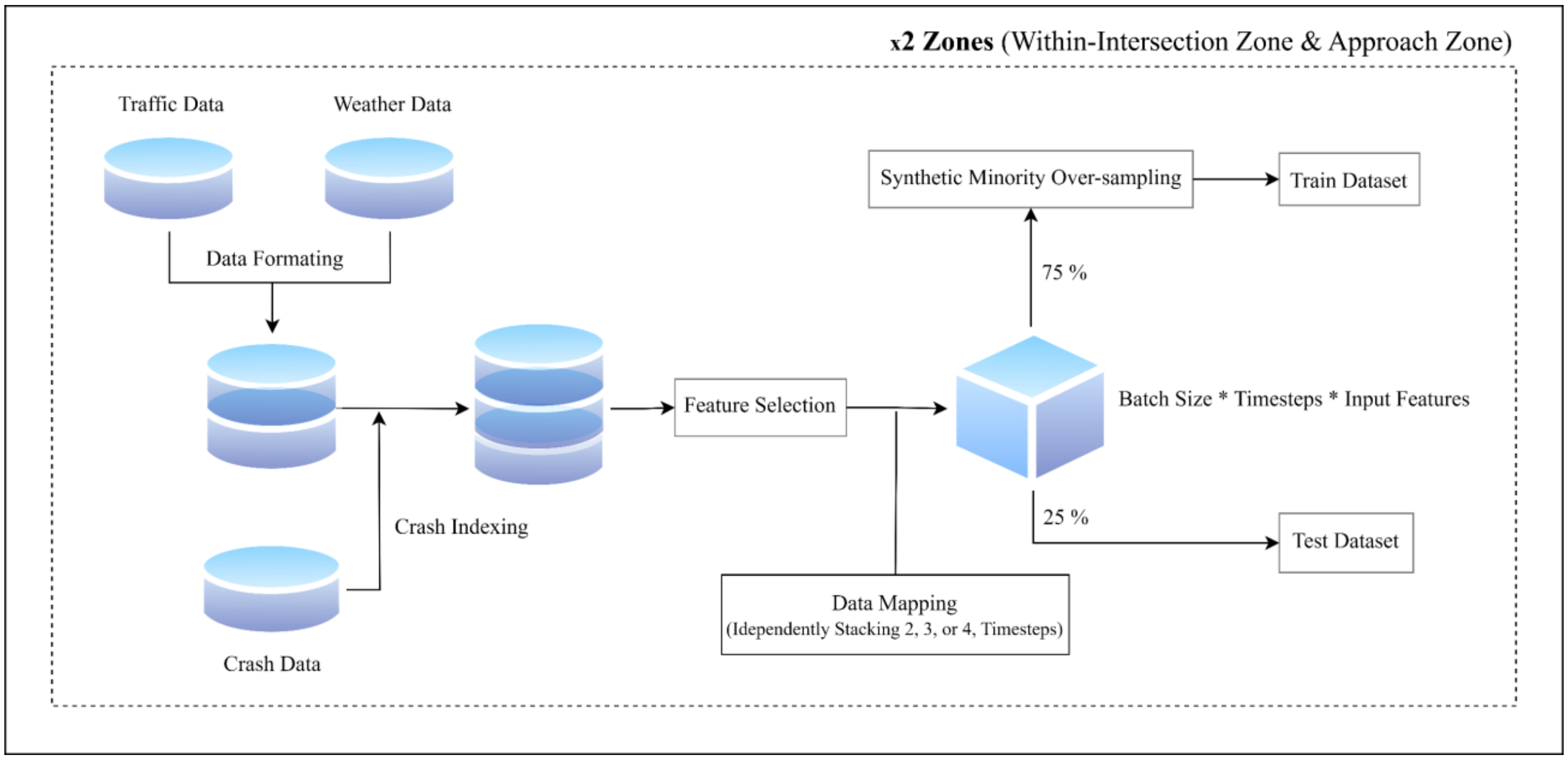

**Figure 6. Data Preparation Pipeline**

The data preparation process presented in **Figure 6** is crucial in real-time modeling of this study. This approach enables the implementation of the real-time model in the context of proactive management.

## RESULTS

In this study, the two-dimensional datasets (with traffic, weather, and crash information) for 'within-intersection' and 'approach' zones were simultaneously mapped to three-dimensional datasets through stacking zone-specific datasets on two, three, and four, timesteps, respectively. This process ensured three stacked datasets for each intersection zone. With the stacked datasets, three zone-specific inTformer models: inTformer (II), inTformer (III), and inTformer (IV), were developed for each intersection zone, where II, III and IV stand for stacking on two, three and four, timesteps, respectively.

### Hyperparameter Tuning

A crucial step to achieving reliable results from the model training is to tune hyperparameters and select appropriate optimization functions. Hence, in this study, all the inTformer models were tuned from a pool of hyperparameters and optimization functions prior to model training. The best set of parameters for the trained inTformer models for each intersection zone is presented in **Table 6**.

**Table 6. Best Parameters from Hyperparameter Tuning**

| Hyperparameters | Zones | |
|---|---|---|
| | **Within-Intersection** | **Approach** |
| | **inTformers (II, III, and IV)** | **inTformers (II, III, and IV)** |
| Learning Rate | 0.00001, 0.0001, 0.0001 | 0.00001, 0.00001, 0.0001 |
| Batch Size | 500, 1000, 1000 | 1000, 1000, 1000 |
| Epoch Number | 50, 50, 100 | 50, 100, 100 |
| No. of Heads | 5, 5, 5 | 5, 5, 5 |
| No. of Encoders | 3, 3, 3 | 3, 3, 4 |
| Optimization Function | Adam, Adam, Adam | Adam, Adam, Adam |

*Here, II, III and IV stand for stacking on two, three and four, timesteps, respectively.*

### Model Evaluation

The performance of the trained models was evaluated on the test datasets obtained from splitting the mapped datasets. The sensitivity and false alarm rate of all the developed inTformer models are presented in **Table 7**.

**Table 7. Experiment Results of Zone-Specific inTformer**

| Zones | Models | Performance Scores | |
|---|---|---|---|
| | | **Sensitivity** | **False Alarm Rate** |
| Within-Intersection | inTformer (II) | **0.73** | **0.36** |
| | inTformer (III) | 0.68 | 0.38 |
| | inTformer (IV) | 0.65 | 0.44 |
| Approach | inTformer (II) | **0.70** | **0.36** |
| | inTformer (III) | 0.67 | 0.39 |
| | inTformer (IV) | 0.65 | 0.43 |

*Here, II, III and IV stand for stacking on two, three and four, timesteps, respectively.*

**Table 7** depicts that the best model for 'within-intersection' zone is inTformer (II) model i.e., inTformer model trained on dataset created by stacking the two-dimensional datasets (with traffic, weather, and crash information) of the 'within-intersection' zone on *two* timesteps. For the 'approach' zone, the inTformer (II) stands out as the top-performing model as well. This model was trained on a dataset created by stacking dataset of the 'approach' zone on *two* timesteps. While the inTformer (II) models for the 'within-intersection' and 'approach' zone yielded decent sensitivity scores (0.73 and 0.70, respectively), they demonstrated relatively poor performance in terms of false alarm rates. This outcome can be attributed to two likely factors. First, false alarm rate has a direct relation with market penetration rate of connected vehicles: as the number of connected vehicles increases in the market, false alarm rate tends to decrease (Islam & Abdel-Aty, 2023). In this study, connected vehicles sourcing traffic data had very low market penetration rate (almost 8%) at the time of data collection. As a result, all the inTformer models trained on connected vehicle's traffic data performed poorly in terms of false alarm rate. Nevertheless, if more connected vehicles penetrate the market, which is a highly expected scenario in the near future, the performance of the inTformer models in terms of false alarm rate as well as sensitivity are expected to improve. Second, the models' inability to capture substantial variations in data distribution also affects the false alarm rates. Given that the traffic data used in this study were aggregated at 15-minute intervals, the inTformer models may have struggled to capture the finer details of traffic flow patterns, resulting in a high false alarm rate. A potential solution to decrease the false alarm rate could be to train the inTformer models on traffic data collected at shorter intervals, such as 1-5 minutes.

**Model Comparison**

To check the performance of the inTformer model, several established deep neural networks, namely LSTM, CNN, sequential LSTM-CNN, parallel LSTM-CNN, were also trained and tested on the same respective mapped datasets. All the compared models were fine-tuned accordingly. **Figure 7** and **Figure 8** respectively present the comparison results of 'within-intersection' zone's inTformer (II) model, and 'approach' zone's inTformer (II) model, in relation to all the established models trained in this study. The comparison results verify that the inTformer models of both 'within-intersection' and 'approach' zone have higher sensitivity and lower false alarm rate compared to other established models developed in this study. This finding proves the validity of zone-specific assignment of the inTformer model for real-time crash likelihood prediction at intersections.

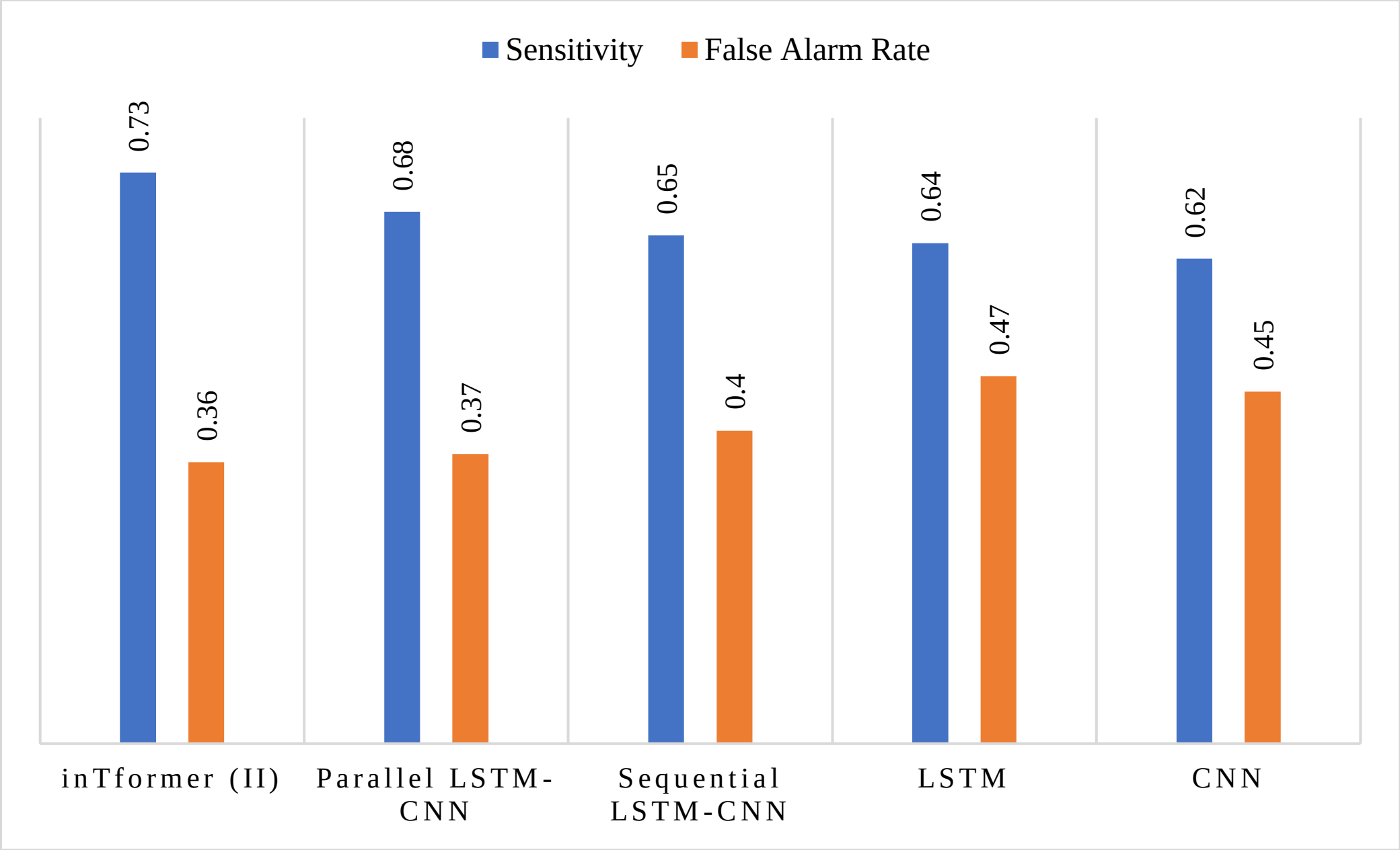

**Figure 7. Comparison Results of 'Within-Intersection' Zone Models**

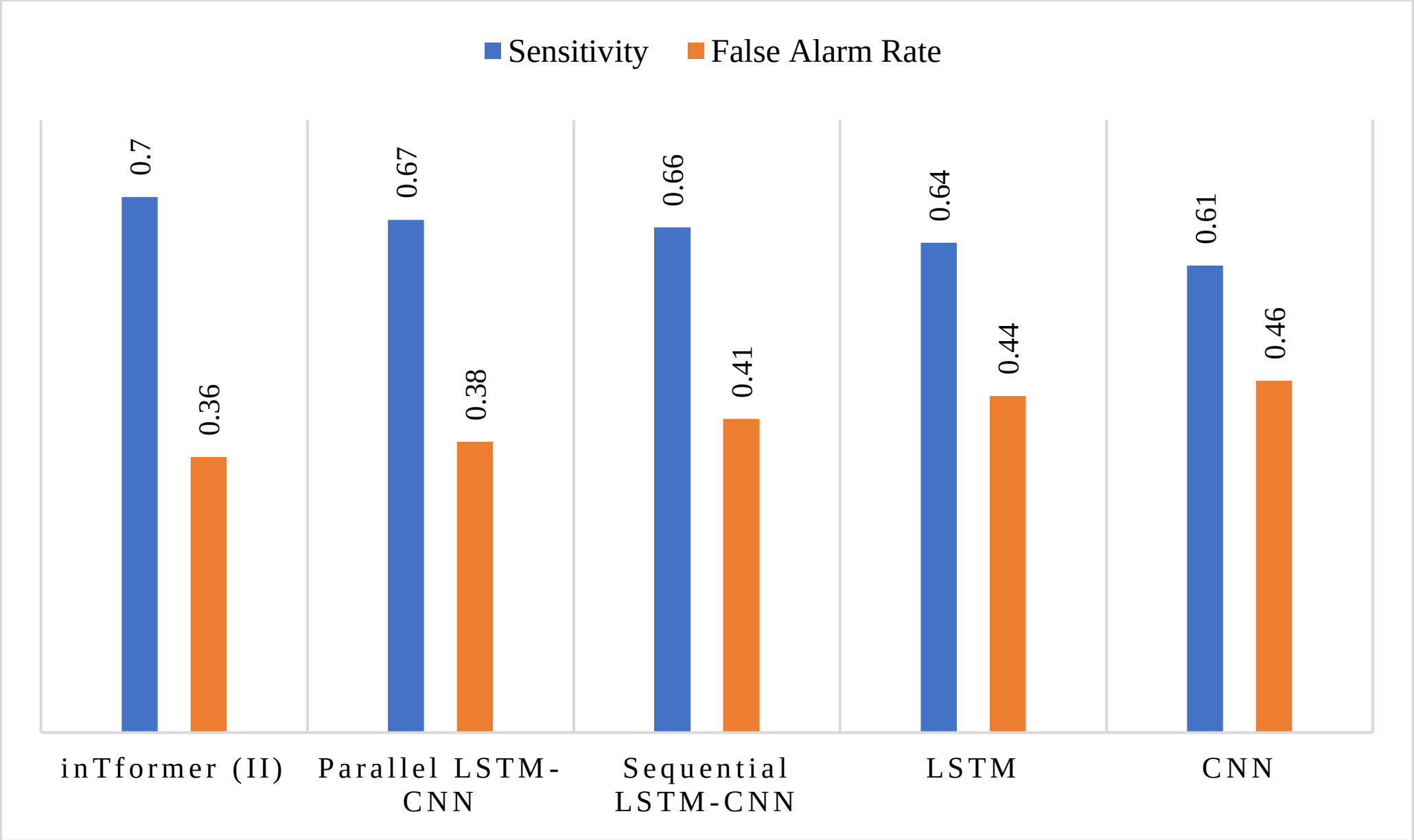

**Figure 8. Comparison Results of 'Approach' Zone Models**

Furthermore, the performance of the proposed inTformer was assessed in relation to previous studies on real-time crash likelihood prediction at intersections. To the best of the author's knowledge, only Yuan et al. (Yuan et al., 2019) applied a deep learning model, namely LSTM, to predict crash likelihood in real-time at intersections. Their analysis was centered on crashes occuring inside intersections (analogous to 'within-intersection' zone of this study). They implied that crashes occuring inside the intersection are more likely to be predicted, corroborating the current study's results that the inTformer (II) for the 'within-intersection' zone is better performing than its counterpart in the 'approach' zone. The comparison of inTformer models of the 'within-intersection,' and 'approach' zone with the proposed LSTM model by Yuan et al. (2019) is presented in **Table 8**.

**Table 8. Model Comparison with Previous Studies**

| Zones | Models | Performance Scores | |
|---|---|---|---|
| | | **Sensitivity** | **False Alarm Rate** |
| Within-Intersection | inTformer (II) | 0.73 | 0.36 |
| Approach | inTformer (II) | 0.70 | 0.36 |
| Within-Intersection | Yuan et al. (2019) | 0.61 | 0.39 |

## Model Interpretation

The proposed inTformer architecture requisites three-dimensional (batch_size × sequence_length (timesteps) × input_feature) sequential data as input. As such, in this study, the better performing inTformer (II) models for the 'within-intersection' and 'approach' zone were built on three-dimensional datasets created by stacking two-dimensional datasets on *two* timesteps. The stacking on *two* timesteps signifies the joint use of features from two successive 15-minute intervals for crash likelihood prediction. Additionally, the stacking also highlights that the features of the successive timesteps have their individual impact on the occurrence of crashes. Apprehending this fact, the game theoretic SHAP method was applied to the inTformer (II) models of the 'within-intersection' and 'approach' zone to identify the importance and contribution of each explanatory feature from the two successive timesteps on the occurrence of crashes.

In general, summary plots with SHAP values are highly effective in expressing the impact of features on outcomes. These plots can not only provide information on the most important features on a global scale, but can also provide ideas on the size, distribution, and direction of the features' contribution on a local scale. The summary plots highlighting the contributions of the top ten features for each timestep in the zone-specific inTformer (II) models are presented in **Figures 9-10**. The features in the figures are ordered by importance. Red dots signify high feature values, while blue dots represent low values. The vertical line positioned at 0.0 distinguishes between positive predictions (representing a crash event) and negative predictions (representing a non-crash event).

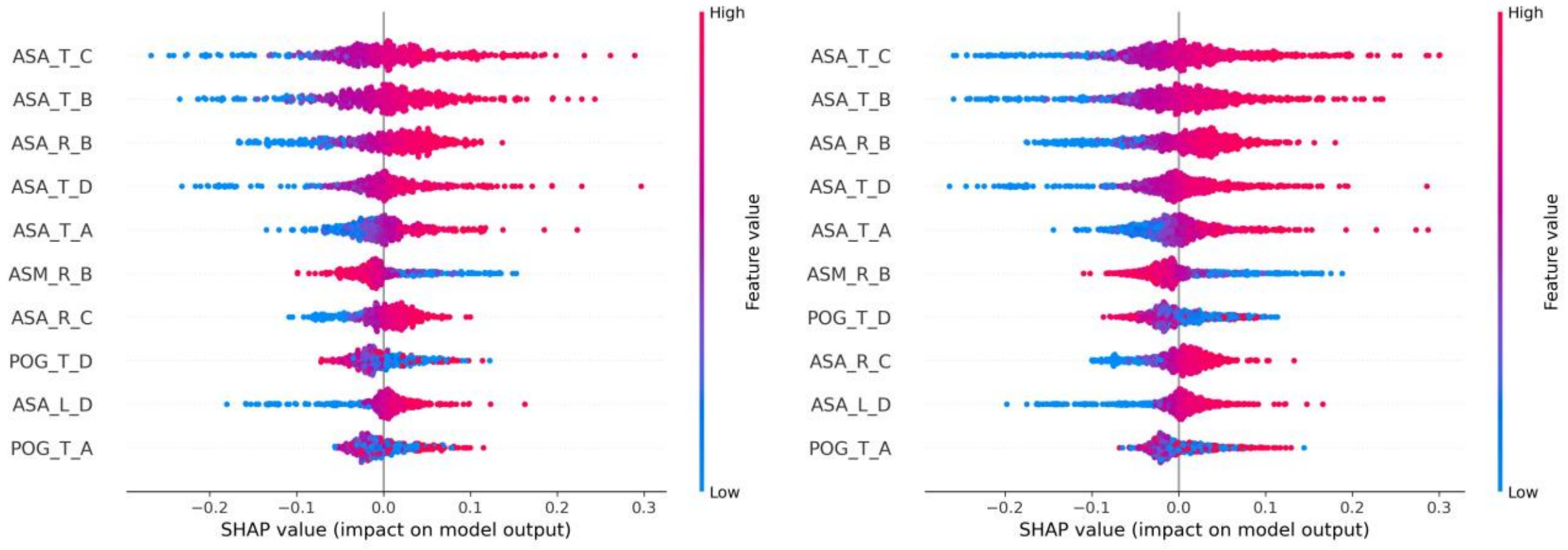

(a) (b)

**Figure 9. Summary Plot of Top Ten Features Impacting Crash in the 'Within-Intersection' Zone, (a) 0-15 Minutes Before a Crash, *and* (b) 15-30 Minutes Before a Crash**

**Figure 9** reveals that, on a global scale, the most important feature influencing crash occurrence at intersection in the 'within-intersection' zone, both withing 0-15 mins (**Figure 9 (a)**) and 15-30 mins (**Figure 9 (b)**) before a crash, is *ASA_T_C*, which represents the Average Speed of Through Vehicles at Approach 'C' (refer to **Figure 4**). The same figures imply that, on a local scale, high value of *ASA_T_C* (represented by red dots) yields high SHAP value (positive predictions), thus increasing likelihood of crashes in the 'within-intersection' zone. In contrast, an increase in the value of *POG_T_D*, which stands for the Percentage of Through Vehicles Arrived on Green at Approach 'D,' appears to reduce the likelihood of crashes (negative predictions) in the 'within-intersection' zone. The rest of the features in **Figures 9(a)** and 9**(b)** are self-explanatory.

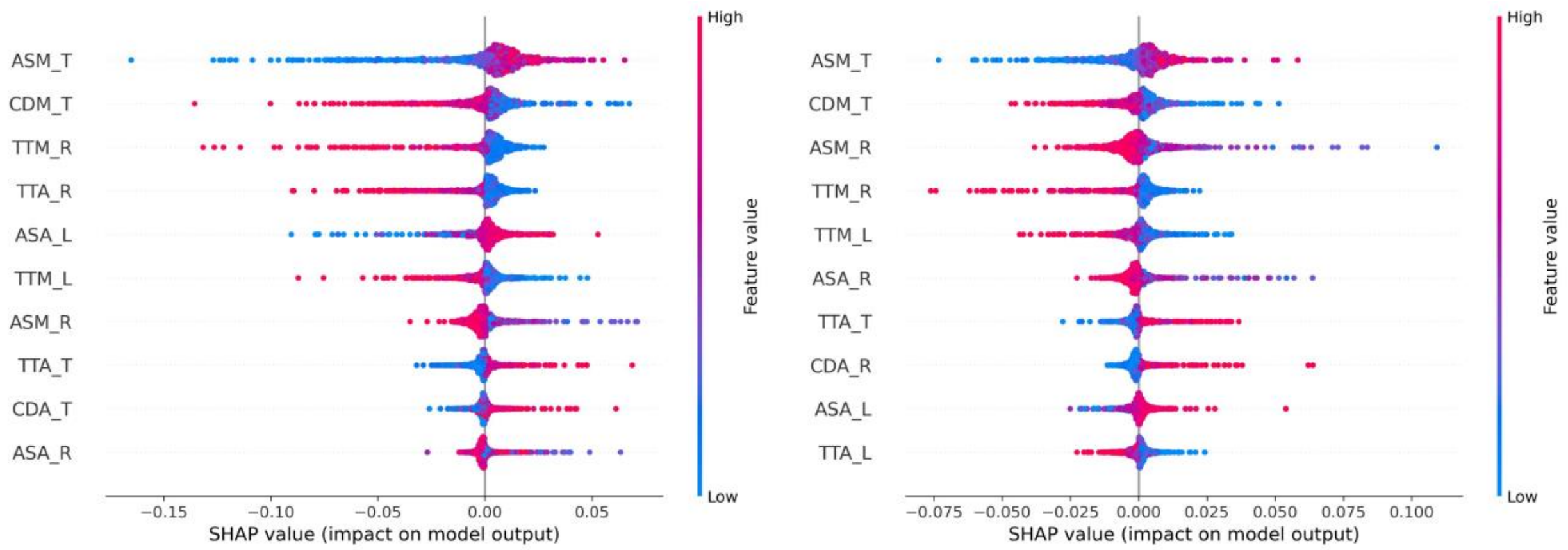

(a) (b)

**Figure 10. Summary Plot of Top Ten Features Impacting Crash in the 'Approach' Zone, (a) 0-15 Minutes Before a Crash, *and* (b) 15-30 Minutes Before a Crash**

In the 'approach' zone, the most important feature, on global scale, was identified to be *ASM_T* (Maximum Speed of Through Vehicles) as highlighted in both **Figures 10(a)** and **10(b)**. A rise in the *ASM_T* value, denoted by red dots, increased the chances of crashes in the 'approach' zone. Similarly, rise in values of *ASA_L* (Average Speed of Left Turning Vehicles), *TTA_T* (Average Travel Time of Through Vehicles), and *CDA_T* (Average Control Delay of Through Vehicles) also increased the likelihood of crashes in the 'approach' zone. Conversely, the increase in the value of all other remaining features in **Figures 10(a)** and **10(b)**, reduced the likelihood of crashes in the 'approach' zone.

## CONCLUSIONS

The traffic operation mechanism at intersections is comparatively more intricate than roadway segments such as freeways and arterials. This intricacy often results in complex crash occurrence patterns, making it challenging to develop a robust model that can predict the likelihood of crashes at intersections. In an effort to address this issue, this study developed zone-specific models for predicting crash likelihood at intersections. To be specific, the study divided the intersection into two different zones: the within-intersection and the approach zone, reflecting the patterns of traffic flow at intersections. Afterwards, separate models to predict the likelihood of crashes were developed for each zone. At methodological level, the authors modified the original Transformer algorithm and proposed the inTformer algorithm for developing zone-specific crash likelihood prediction models. The data for training the inTformer model comprised of crash data, connected vehicle data (representing traffic operation dynamics at intersections), and weather data. All the data were independently processed and prepared, taking into account the defined zone-level traffic operation and crash incidents, prior to their application in the final model training.

After data preparation, a total of six real-time crash likelihood prediction models: three for 'within-intersection' zone and three for 'approach' zone, were developed. For the 'within-intersection' zone, the inTformer (II) model outperformed all models with a sensitivity of 73% and false alarm rate of 36%. This model was built by training inTformer algorithm on the dataset generated from stacking 'within-intersection' zone dataset stacked on two timesteps. Similarly, for the 'approach' zone, the inTformer (II) model was also the top performer, with a sensitivity of 70% and a false alarm rate of 36%. The efficacy of the developed zone-specific inTformer (II) models were also evaluated using several existing deep neural networks such as LSTM, CNN, sequential-LSTM, and parallel-LSTM models. In all scenarios, the inTformer models demonstrated superior performance compared to the existing deep learning models. Furthermore, a comparison with the previously developed real-time models for intersection crash prediction further confirmed the superiority of the proposed inTformer model (Yuan et al., 2019).

Furthermore, to ensure explainability of the proposed inTformer models, SHAP technique was employed on the inTformer (II) models of both 'within-intersection' and 'approach' zones, quantifying the impact of explanatory features on the likelihood of crashes at intersections. In the 'within-intersection' zone, features like average approach speed, maximum approach speed, and the percentage of vehicles on green were the critical predictors of crash likelihood. Meanwhile, in the 'approach' zone, the determining features included average and maximum approach speeds, average and maximum control delays, as well as average and maximum travel times.

In summary, this paper succeeds in verifying the viability of real-time crash likelihood prediction at intersections using the proposed inTformer. The results of the inTformer model indicate its suitability for the implementation of an advanced traffic management system that has the potential to reduce crashes. Nevertheless, the current research has several unexplored directions. Firstly, embedding extra layers in the inTformer architecture or tuning the inTformer on additional hyperparameter combinations. Secondly, resampling techniques other than SMOTE or data augmentation can be implemented to check out the possibility of enhanced model performance. Thirdly, driver characteristics extracted from connected vehicles, such as hard braking, hard acceleration, etc., can be incorporated into model training, since it has been extensively exhibited that crash events are highly prompted by driver characteristics and driving behavior prior to crash occurrence. Finally, Signal Analytics' connected vehicle data had no information on Signal Phase and Timing (SPaT). As SPaT has direct relation with traffic flow parameters at intersections, including this data type with Signal Analytics' connected vehicle data can potentially enhance the prediction model's performance.

## AUTHOR CONTRIBUTIONS

The authors confirm contribution to the paper as follows: study conception and design: Anik. BMTH, Islam. Z, Aty. MA; data collection: Anik. BMTH, Islam. Z, Aty. MA; analysis and interpretation of results: Anik. BMTH, Islam. Z, Aty. MA; draft manuscript preparation: Anik. BMTH, Islam. Z, Aty. MA. All authors reviewed the results and approved the final version of the manuscript.